\documentclass[review]{elsarticle}

\usepackage{lineno,hyperref}
\usepackage{comment}
\usepackage{amsmath}
\usepackage{amsthm}
\usepackage{amsfonts}
\usepackage{algorithm}
\usepackage{graphicx}
\usepackage{subfigure}
\usepackage[noend]{algpseudocode}

\usepackage{amssymb}
\usepackage{multirow}
\usepackage{array}
\usepackage{enumitem}
\usepackage{colortbl}
\usepackage{makecell}
\usepackage[table]{xcolor}
\newcommand{\highlight}[1]{\textcolor{black}{#1}}

\modulolinenumbers[5]

\journal{Journal of Knowledge-Based Systems}

\begin{document}

\begin{frontmatter}

\title{Dialogue Summarization with Supporting Utterance Flow Modeling and Fact Regularization}

\author[CUHK_CSE]{Wang Chen\corref{mycorrespondingauthor}}
\ead{wchen@cse.cuhk.edu.hk}

\author[AI_Lab]{Piji Li}
\ead{lipiji.pz@gmail.com}

\author[Macau_U]{Hou Pong Chan}
\ead{hpchan@um.edu.mo}

\author[CUHK_CSE]{Irwin King}
\ead{king@cse.cuhk.edu.hk}

\address[CUHK_CSE]{Department of Computer Science and Engineering, The Chinese University of Hong Kong, Shatin, N.T., Hong Kong SAR, China}
\address[Macau_U]{Department of Computer and Information Science, University of Macau, Macau SAR, China.}
\address[AI_Lab]{Tencent AI Lab, Shenzhen, China}

\cortext[mycorrespondingauthor]{Corresponding author}




\begin{abstract}
Dialogue summarization aims to generate a summary that indicates the key points of a given dialogue. In this work, we propose an end-to-end neural model for dialogue summarization with two novel modules, namely, the \emph{supporting utterance flow modeling module} and the \emph{fact regularization module}. The supporting utterance flow modeling helps to generate a coherent summary by smoothly shifting the focus from the former utterances to the later ones. The fact regularization encourages the generated summary to be factually consistent with the ground-truth summary during model training, which helps to improve the factual correctness of the generated summary in inference time. 
Furthermore, we also introduce a new benchmark dataset for dialogue summarization. Extensive experiments on both existing and newly-introduced datasets demonstrate the effectiveness of our model. 
\end{abstract}

\begin{keyword}
\texttt{Dialogue Summarization}\sep Text Summarization \sep Text Generation
\end{keyword}

\end{frontmatter}


\section{Introduction}

\begin{figure}[t]
\centering
\includegraphics[width=0.6\columnwidth]{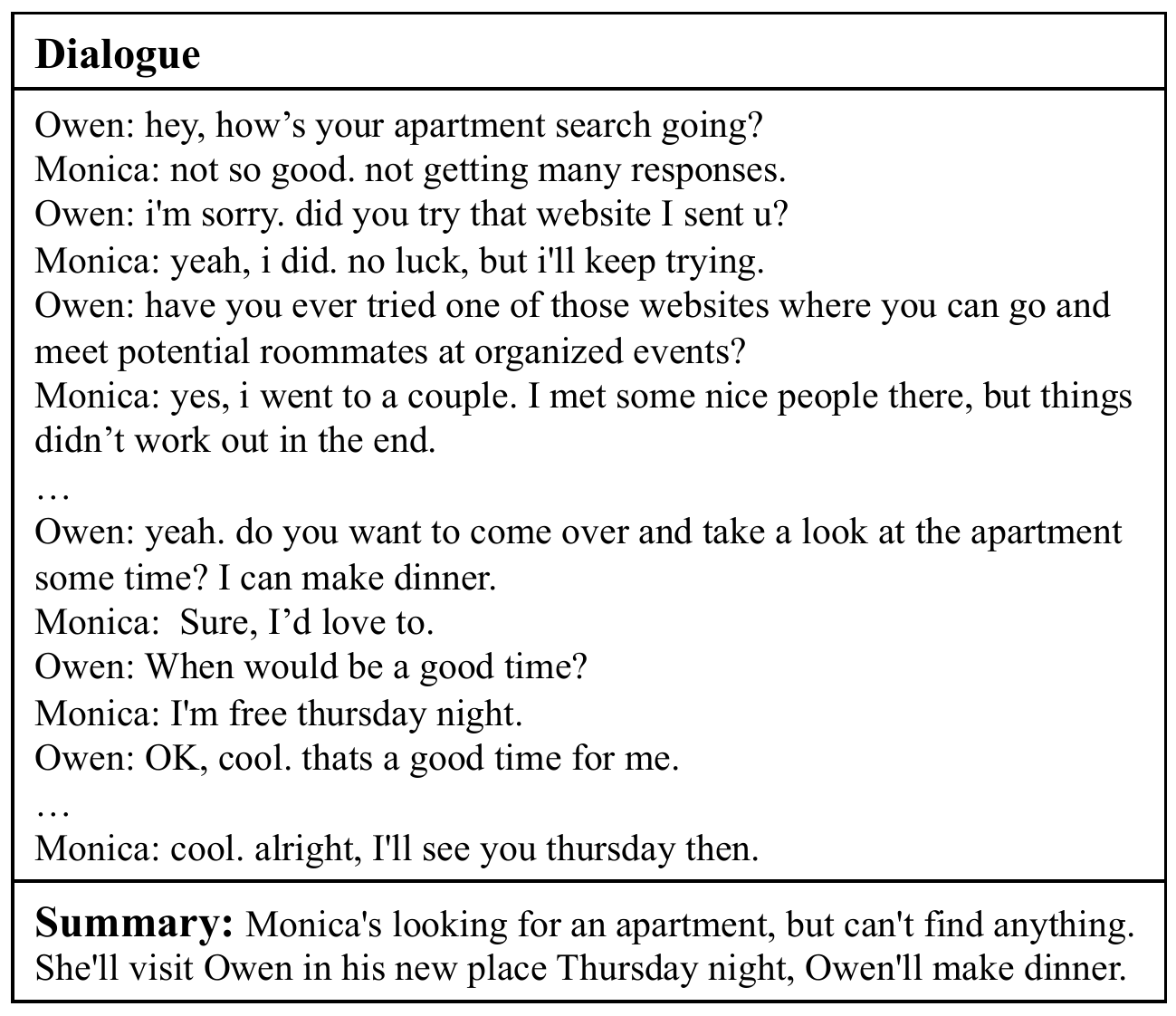}
\caption{An example of dialogue summarization. \highlight{Each complete dialogue line here is regarded as one utterance of the dialogue}.}
\label{figure: case_study_intro}
\vspace{-0.1in}
\end{figure}

The task of dialogue summarization focuses on summarizing the key information of a given dialogue into a short and concise summary as shown in Figure~\ref{figure: case_study_intro}. The summary can assist humans to quickly acquire the key points without reading the entire dialogue which may be long and twisted~\cite{DBLP:conf/kdd/LiuWXLY19}. Therefore, dialogue summarization is a significant research problem and has been widely applied in various applications, such as summarizing meetings~\cite{DBLP:conf/acl/LiZJR19}, medical conversations~\cite{DBLP:conf/asru/LiuNGAC19}, and customer service dialogues~\cite{DBLP:conf/kdd/LiuWXLY19}. 

The lack of suitable benchmark datasets is a long-standing problem in the area of dialogue summarization~\cite{DBLP:conf/mlmi/CarlettaABFGHKKKKLLLMPRW05,DBLP:conf/slt/GooC18}. 
To address this problem, Gliwa et al.~\cite{gliwa-etal-2019-samsum} published the SAMSum dataset, which is a large-scale dialogue summarization dataset with human-annotated summaries. Moreover, they also applied several state-of-the-art news summarization models~\cite{see-etal-2017-get,chen-bansal-2018-fast, DBLP:conf/iclr/WuFBDA19} to solve the dialogue summarization problem and achieved promising results, but they do not propose any new and specific models to this task. However, we observe that dialogue summarization has two important characteristics that are different from news summarization. Simply applying news summarization models on dialogue summarization may lead to a suboptimal solution. 

\begin{table}[t]
\centering
\resizebox{0.6\columnwidth}{!}{
\begin{tabular}{ | c || c | c | c || c | c | c |}
\hline
\multicolumn{1}{|c||}{\multirow{2}{*}{\textbf{\makecell{Position \\ Range}}}} & \multicolumn{3}{c||}{\textbf{SAMSum}} & \multicolumn{3}{c|}{\textbf{CNN/DailyMail}} \\ \cline{2-7}
\multicolumn{1}{|c||}{}   & S1    & S2    & S3  & S1   & S2  & S3 \\
\hline
(0.0, 0.1)
& 0.15 & 0.03 & 0.02
& \cellcolor[HTML]{C0C0C0} \textbf{0.34} & \cellcolor[HTML]{C0C0C0}\textbf{0.19} & 0.11 \\

[0.1, 0.2)
& \cellcolor[HTML]{C0C0C0}\textbf{0.20} & 0.07 & 0.04
& \cellcolor[HTML]{C0C0C0}\textbf{0.26} & \cellcolor[HTML]{C0C0C0}\textbf{0.23} & \cellcolor[HTML]{C0C0C0}\textbf{0.17} \\

[0.2, 0.3)
& \cellcolor[HTML]{C0C0C0}\textbf{0.19} & 0.10 & 0.06
& 0.14 & 0.16 & \cellcolor[HTML]{C0C0C0}\textbf{0.15} \\

[0.3, 0.4)
& 0.12 & 0.11 & 0.07
& 0.07 & 0.11 & 0.12\\

[0.4, 0.5)
& 0.06 & 0.10 & 0.08
& 0.05 & 0.08 & 0.10\\

[0.5, 0.6)
& 0.08 & \cellcolor[HTML]{C0C0C0}\textbf{0.13} & 0.12
& 0.04 & 0.07 & 0.09\\

[0.6, 0.7)
& 0.07 & \cellcolor[HTML]{C0C0C0}\textbf{0.13} & 0.14
& 0.03 & 0.05 & 0.08\\

[0.7, 0.8)
& 0.05 & 0.10 & 0.13
& 0.02 & 0.04 & 0.06\\

[0.8, 0.9)
& 0.04 & 0.12 & \cellcolor[HTML]{C0C0C0}\textbf{0.17}
& 0.02 & 0.03 & 0.06\\

[0.9, 1.0]
& 0.04 & 0.11 & \cellcolor[HTML]{C0C0C0}\textbf{0.17}
& 0.03 & 0.04 & 0.06 \\

\hline
Sum
& 1 & 1 & 1
& 1 & 1 & 1\\
\hline
\end{tabular}
}
\caption{The position distributions of supporting utterances (sentences) of the first three summary sentences in SAMSum (CNN/DailyMail) dataset. ``S$i$'' means the $i$-th summary sentence. 
The first column indicates the relative utterance (sentence) position range. For example, ``[0.1, 0.2)'' means the range of the first 10\%-20\% utterances (sentences). The top-2 ratios for each summary sentence are bold.
}
\label{table: SUF_distribution}
\vspace{-0.1in}
\end{table}

First, a news summary usually focuses on the first few sentences of a news article~\cite{see-etal-2017-get}, while a dialogue summary smoothly transits its focus from the beginning to the end of the dialogue. We define the supporting utterances (sentences) of a summary sentence as the most informative ones which obtain the highest Jaccard similarities with the summary sentence after removing all the stop-words. We then illustrate the position distribution of supporting utterances (sentences) of the first three summary sentences on the dialogue (news) summarization benchmark in Table~\ref{table: SUF_distribution}. It is observed that most of the supporting sentences of the first three summary sentences in the news summarization benchmark CNN/DailyMail~\cite{DBLP:conf/nips/HermannKGEKSB15,nallapati-etal-2016-abstractive} appear in the 0\%-30\% part of the source news input. On the other hand, a dialogue summary in SAMSum smoothly transits the focus from the start to the end of the dialogue. Specifically, the supporting utterances of the first summary sentence (i.e., S1) are mainly distributed on the 10\%-30\% part of a dialogue, but the supporting utterances of the third summary sentence (i.e., S3) mainly appear in the 80\%-100\% part. 
Former (later) summary sentences focus on the former (later) dialogue utterances. We call such a feature as \emph{supporting utterance flow}.

Second, dialogue summaries usually contain more frequent fact triplets with a subject-verb-object structure than news summaries. We extract subject-verb-object (SVO) fact triplets from the summaries of both CNN/DailyMail and SAMSum datasets. We find that the summaries of the SAMSum dataset have 0.803 fact triplets per sentence, whereas the summaries of CNN/DailyMail dataset only have 0.697 fact triplets per sentence. Thus, it is desirable to exploit the information of fact triplets in the ground-truth dialogue summaries to help the model learn to predict dialogue summaries with rich and accurate facts.

Motivated by the above observations, we propose a novel dialogue summarization model that explicitly incorporates the supporting utterance flow and subject-verb-object fact triplets in dialogue summaries. Our basic model employs a hierarchical encoder to encode the dialogue and a decoder with a hierarchical attention mechanism to generate the summary. For modeling the supporting utterance flow feature, we propose a supporting utterance flow modeling (SUFM) module. This module consists of an SUFM embedding and an SUFM loss. The SUFM embedding injects the correlation between the utterance position and the summary token position into our model. The SUFM loss encourages our model to smoothly transit the focus from the start to the end of the dialogue and ignore the utterances that have been summarized by formerly generated summary sentences. 
To incorporate the information of fact triplets, we propose a fact regularization (FR) module.
In this module, SVO fact triplets are first extracted from ground-truth summaries by a fact triplet extractor. Then we introduce an FR loss into our training objective, which encourages the generated summary to be factually consistent with the ground-truth summary. Hence, the FR loss helps to boost the factual accuracy of the summaries generated by our model.

Furthermore, we also \highlight{generate a new dataset from an existing video-based dialogue corpus for dialogue summarization to enrich the benchmarks in this area}. Comprehensive experiments are conducted on SAMSum and the newly-introduced dataset. The results show that our model outperforms multiple state-of-the-art news summarization models on both automatic and human evaluations. Besides, the ablation study indicates the effectiveness of our SUFM and FR modules. We also find that our model can learn the smooth transition of the supporting utterance flow and generate more important facts under the partially-matched metric.

In summary, our main contributions in dialogue summarization are as follows: (1) a novel framework which explicitly models the supporting utterance flow and incorporates the information of fact triplets to improve factual correctness; (2) a new benchmark dataset which can relieve the lack of benchmark datasets in this area and can be utilized by the research community for further studies; (3) the better performance than directly applied state-of-the-art news summarization models on the two dialogue summarization benchmarks. 

\section{Related Work}
Most of existing work on dialogue summarization focuses on summarizing meeting dialogues using the AMI meeting corpus~\cite{DBLP:conf/mlmi/CarlettaABFGHKKKKLLLMPRW05}. 
Early literature of dialogue summarization proposed different extractive methods~\cite{DBLP:conf/slt/XieLL08,DBLP:conf/interspeech/GargFRH09}, template-based generation methods~\cite{DBLP:conf/acl/WangC13,DBLP:conf/asru/XieHFL09,oya-etal-2014-template}, and graph-based generation methods~\cite{DBLP:conf/enlg/MehdadCTN13,shang-etal-2018-unsupervised}. 
Recently, neural generative models~\cite{li-etal-2019-keep,DBLP:conf/www/ZhaoPFLLY19} were proposed and they achieved state-of-the-art performance. However, the AMI meeting corpus only has 140 meeting samples, which is too small to adequately reflect the data distribution.

To address this limitation, various work developed new and large datasets for dialogue summarization like~\cite{DBLP:conf/slt/GooC18,DBLP:journals/corr/abs-1811-00185,DBLP:journals/corr/abs-1910-00825}.  Nevertheless, the summaries from these datasets are either too short and general to show all the key points of the dialogue or not human-annotated, thus lacking fluency and coherence. Although time-consuming and costly, human-annotating is still the best way to annotate the summary of dialogue when preparing the benchmark dataset. Therefore, Liu et al.~\cite{DBLP:conf/kdd/LiuWXLY19} collected a large-scale dialogue summarization dataset for customer service by human-annotating, but this dataset is not public.

Lately, Gliwa et al.~\cite{gliwa-etal-2019-samsum} released a public, large-scale, and human-annotated dialogue summarization dataset SAMSum. The summaries of the messenger-like dialogues are annotated by language experts. Gliwa et al.~\cite{gliwa-etal-2019-samsum} evaluated plenty of news summarization models on this dataset, but they did not propose any new model specific to dialogue summarization. Feng et al.~\cite{DBLP:journals/corr/abs-2010-10044} incorporated the commonsense knowledge when summarizing dialogues. Chen and Yang~\cite{chen-yang-2020-multi} considered the conversation structure and designed a multi-view sequence-to-sequence model for this task. Zhao et al.~\cite{zhao-etal-2020-improving} utilized the graph structures and topic words to improve the performance. 
However, these previous work ignored the specific features of dialogue summarization like supporting utterance flow and a higher frequency of SVO fact triplets in summaries. Thus, in this work, we propose an SUFM module which is inspired by Gao et al.~\cite{gao-etal-2019-interconnected} to explicitly model supporting utterance flow and an FR module to utilize the information of fact triplets and improve the factual accuracy of \highlight{generated summaries}. 
Moreover, we also introduce a new human-annotated dataset to mitigate the lack of high-quality benchmarks in this area.

\section{Our Model}
We formally define the dialogue summarization problem as follows. Given a dialogue context $\mathbf{\tilde{X}}$, the goal is to generate the summary $\mathbf{y}$ of the dialogue. The dialogue context is a sequence of utterances, i.e., $\mathbf{\tilde{X}}=[\mathbf{x}_1,..., \mathbf{x}_i, ..., \mathbf{x}_{L_{\mathbf{\tilde{X}}}}]$ where $\mathbf{x}_i$ is the $i$-th utterance and $L_{\mathbf{\tilde{X}}}$ is the number of utterances of dialogue $\mathbf{\tilde{X}}$. Each utterance $\mathbf{x}_i$ is a sequence of tokens, i.e., $\mathbf{x}_i = [x_{i,1}, ..., x_{i,j}, ..., x_{i,L_{\mathbf{x}_{i}}}]$ where $x_{i,j}$ is the $j$-th token of $\mathbf{x}_i$ and $L_{\mathbf{x}_{i}}$ is the number of tokens of $\mathbf{x}_i$. Similarly, the summary $\mathbf{y}= [y_1, ..., y_t, ..., y_{L_{\mathbf{y}}}]$ is also a sequence with $L_{\mathbf{y}}$ tokens.

\begin{figure}[t]
  \centering
  \includegraphics[width=0.7\columnwidth]{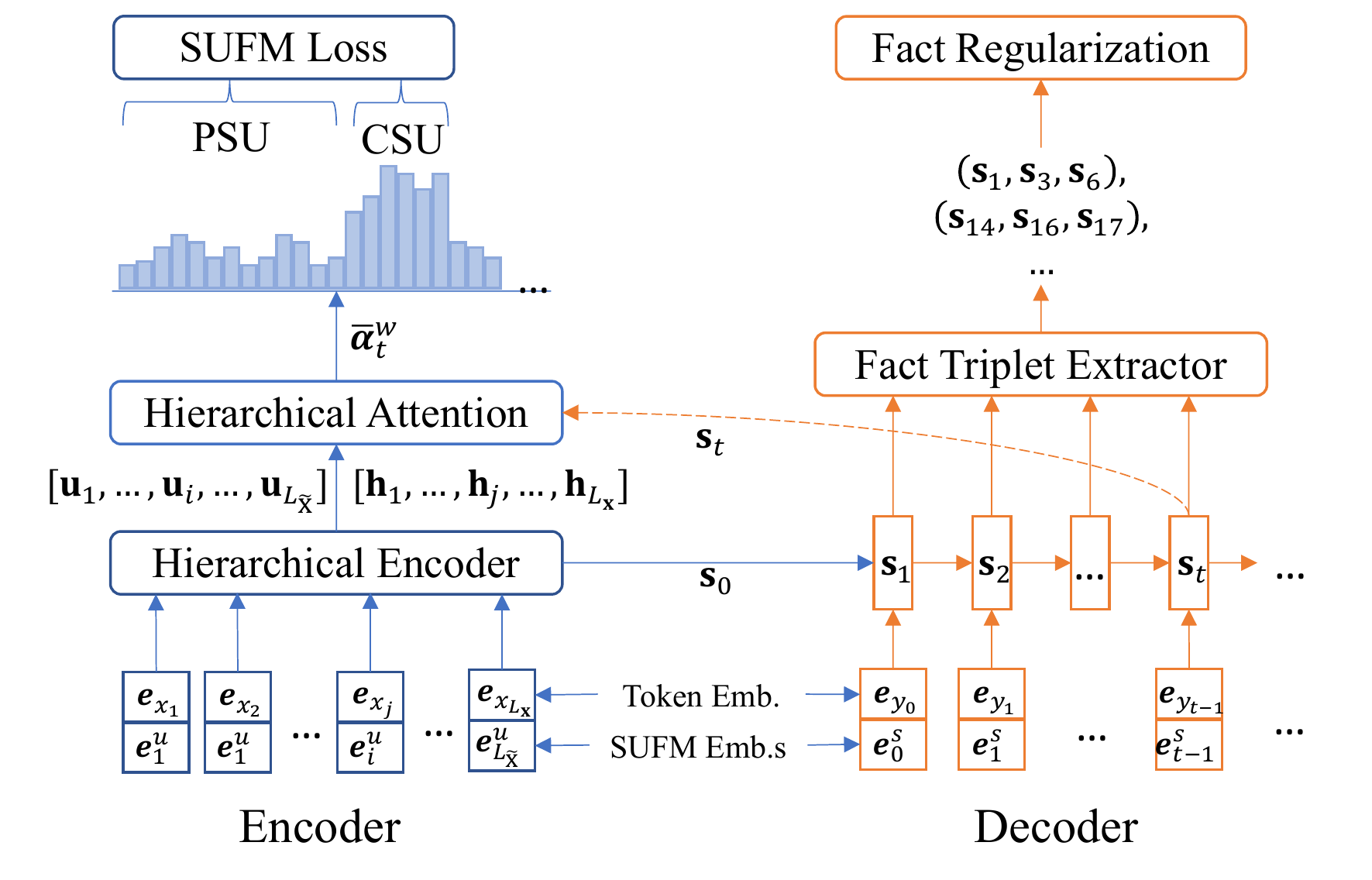}
  \caption{Overall framework of our model. The ``SUFM Loss'' and ``SUFM Emb.s'' compose our SUFM module. The ``Fact Triplet Extractor'' and ``Fact Regularization'' compose of our fact regularization module. $\mathbf{e}^u_i$ is the position embedding of the $i$-th dialogue utterance. $\mathbf{e}^s_t$ is the position embedding of the $t$-th summary token. $\mathbf{u}_i$ is the hidden representation of the $i$-th utterance. $\mathbf{h}_j$ is the hidden representation of the $j$-th token of the dialogue. $\mathbf{\bar{\alpha}}^w_t$ is the rescaled word-level attention scores of dialogue tokens. ``PSU'' and ``CSU'' means the supporting utterances of the previous summary sentences and the current summary sentence respectively.
  }
  \label{figure: model_framework}
\end{figure}

Our full model is illustrated in Figure~\ref{figure: model_framework}. Our model consists of three parts: (1) the basic model with a hierarchical encoder and a decoder with hierarchical attention; (2) the supporting utterance flow modeling (SUFM) module; (3) the fact regularization (FR) module. 

Our SUFM module is composed of a SUFM embedding and a SUFM loss aiming at helping the basic model smoothly transit its focus from the beginning to the end of the dialogue. \highlight{The SUFM embedding is denoted as ``SUFM Emb.s'' in Figure~\ref{figure: model_framework} to distinguish it from the regular token embedding.} It is incorporated into the encoding and decoding process of the basic model for providing a word-level alignment between the generated summary tokens and the dialogue utterances. The SUFM loss is utilized during the training stage to provide sentence-level alignment between the summary sentences and the dialogue utterances.
The FR module is a regularization loss incorporated in the training stage to help our model improve factual correctness when generating summaries. We introduce the basic model first and then the SUFM and FR modules.

\subsection{Basic Model}
\subsubsection{Hierarchical Encoder}
The hierarchical encoder converts the dialogue into two-level hidden representations. 
We append a separator token ``$|$'' to each utterance $\mathbf{x}_i$ and denote the new one as $\mathbf{x}^*_i$. Then, we concatenate all the utterances together as the source input $\mathbf{x}$ of our model where $\mathbf{x}=[x_1, ..., x_j, ..., x_{L_{\mathbf{x}}}]$ and $L_{\mathbf{x}} = \sum_i (L_{\mathbf{x}_i}$ + 1). We map each token $x_j$ into a hidden vector $\mathbf{e}_{x_j} \in \mathbb{R}^{d_e}$ \highlight{via a token embedding layer}. Then, we employ a bi-directional Gated Recurrent Unit (GRU)~\cite{DBLP:conf/emnlp/ChoMGBBSB14} encoder layer to learn the word-level representations $[\mathbf{h}_1, ..., \mathbf{h}_j, ..., \mathbf{h}_{L_{\mathbf{x}}}]$ where $\mathbf{h}_j \in \mathbb{R}^d$. \highlight{After that, an average pooling layer with dropout is applied to each utterance and the obtained vectors are input into another bi-directional GRU layer to learn the utterance-level representations $[\mathbf{u}_1, ..., \mathbf{u}_i, ..., \mathbf{u}_{L_{\mathbf{\tilde{X}}}}]$ where $\mathbf{u}_i \in \mathbb{R}^d$. We concatenate the last forward and backward hidden states (i.e., $[\overrightarrow{\mathbf{u}}_{L_{\mathbf{\tilde{X}}}};\overleftarrow{\mathbf{u}}_1] \in \mathbb{R}^d$) as the global representation of the whole dialogue.} The ``$[\cdot ; \cdot]$" denotes concatenation.

\subsubsection{Decoder with Hierarchical Attention}
Based on the learned word-level, utterance-level, and global representations of the dialogue, the decoder generates an output summary $\mathbf{y}=[y_1, ..., y_t, ..., y_{L_{\mathbf{y}}}]$  token by token. We employ a unidirectional GRU layer as the decoder.

First, the decoder updates its hidden state: 
$\mathbf{s}_t = \text{GRU}(\mathbf{e}_{y_{t-1}}, \mathbf{s}_{t-1}) \text{,}$
where $\mathbf{e}_{y_{t-1}} \in \mathbb{R}^{d_e}$ is the embedding vector of $y_{t-1}$ and $\mathbf{s}_t \in \mathbb{R}^d, t=1,...,L_{\mathbf{y}}$. \highlight{$\mathbf{e}_{y_0}$ is the embedding of the start token and $\mathbf{s}_0 = [\overrightarrow{\mathbf{u}}_{L_{\mathbf{\tilde{X}}}};\overleftarrow{\mathbf{u}}_1]$.}

Then, we use a hierarchical attention mechanism to gather dialogue information from two levels. The utterance-level attention is as follows:
\begin{align}
    \beta^u_{t,i} &= (\mathbf{v}^u)^T\text{tanh}(\mathbf{W}^u_1 \mathbf{s}_t + \mathbf{W}^u_2 \mathbf{u}_i + \mathbf{b}^u) \text{,} \\
    \alpha^u_{t,i} &= \exp(\beta^u_{t,i}) \slash \sum_{k=1}^{L_{\mathbf{\tilde{X}}}} \exp(\beta_{t,k}) \text{,} \\
    \mathbf{c}^u_t &= \sum_{i=1}^{L_{\mathbf{\tilde{X}}}} \alpha^u_{t,i} \mathbf{u}_i \text{,}
\end{align}
where the superscript ``$u$'' means utterance level and $\mathbf{v}^u \in \mathbb{R}^d$ is a trainable vector. $\alpha^u_{t,i}$ is the utterance-level attention score of $\mathbf{x}_i$ and $\mathbf{c}^u_t \in \mathbb{R}^d$ is the aggregated utterance-level contextual information. In this paper, $\mathbf{W}$ and $\mathbf{b}$ are used to denote a projection matrix and a bias vector.
\highlight{In the word-level attention, the initial word-level attention score of the $j$-th token (i.e., $\alpha^w_{t,j}$) is computed like $\alpha^u_{t,i}$ but using another set of parameters and replacing $\mathbf{u}_i$ with $\mathbf{h}_j$. Besides, we rescale each $\alpha^w_{t,j}$ by the utterance-level attention score of the utterance that $x_j$ belongs to. After normalization, we obtain the rescaled word-level attention score and denote it as $\bar{\alpha}^w_{t,j}$. The purpose of the rescaling is to provide utterance-level guidance when performing word-level attention. We utilize $\bar{\alpha}^w_{t,j}$ to gather word-level contextual vector $\mathbf{c}^w_t = \sum_{j=1}^{L_{\mathbf{x}}} \bar{\alpha}^w_{t,j} \mathbf{h}_j$. After gathering $\mathbf{c}^w_t$ and $\mathbf{c}^u_t$, we merge them together with the current decoder state via a MLP layer with dropout: $\bar{\mathbf{s}}_t = \text{Dropout}(\mathbf{W}^m[\mathbf{s}_t;\mathbf{c}^u_t;\mathbf{c}^w_t]+\mathbf{b}^m, r)$, where $r$ is the dropout ratio.}

Finally, we utilize the merged state $\bar{\mathbf{s}}_t$ to predict the probability distribution of the current summary token. The copy mechanism~\cite{see-etal-2017-get} is incorporated in this stage: \highlight{$P(y_t) = (1-g_t)P_\mathcal{V}(y_t) + g_t P_\mathcal{X}(y_t)$, where $g_t = \text{sigmoid}(\mathbf{W}^g \bar{\mathbf{s}}_t + b^g) \in \mathbb{R}$ is the copy gate, $P_\mathcal{V}(y_t) = \text{softmax}(\mathbf{W}^{\mathcal{V}} \bar{\mathbf{s}}_t + \mathbf{b}^{\mathcal{V}}) \in \mathbb{R}^{|\mathcal{V}|}$ is the probability distribution over the predefined vocabulary $\mathcal{V}$, $P_\mathcal{X}(y_t) = \sum_{j:x_j=y_t} \bar{\alpha}^w_{t,j} \in \mathbb{R}^{|\mathcal{X}|}$ is the copy probabilities over $\mathcal{X}$ which is a set of all the tokens appeared in the source input $\mathbf{x}$, $P(y_t) \in \mathbb{R}^{|\mathcal{V} \cup \mathcal{X}|}$ is the final predicted probability distribution.  We set $P_\mathcal{V}(y_t)=0$ if $y_t \notin \mathcal{V}$. Similarly, we set $P_\mathcal{X}(y_t)=0$ if $y_t \notin \mathcal{X}$.}

We apply a typical negative log-likelihood loss as the generation loss of our model:
\begin{align}
    l_{G} = - \sum_{t=1}^{L_{\mathbf{y}}} \text{log} P(y_t | y_1, ..., y_{t-1}; \mathbf{x}; \mathcal{F}) \text{,}
\end{align}
where ``G'' means generation. $\mathcal{F}$ is a set of other features that can be incorporated into the model. For our basic model, $\mathcal{F}$ is empty.

\subsection{Supporting Utterance Flow Modeling}
Our supporting utterance flow modeling (SUFM) module aims at explicitly encouraging our model to learn a smooth focus transition from the beginning to the end of the dialogue when generating the summary. Our SUFM module consists of an SUFM embedding and an SUFM loss.

\smallskip
\noindent \textbf{SUFM Embedding.} We utilize the SUFM embedding to indicate the correlation between the utterance position and the summary token position. It consists of two embedding layers, i.e., an utterance position embedding layer for the dialogue input and a position embedding layer for the summary output. 

For the input token $x_j$, the utterance position embedding layer maps the index of the utterance that the token $x_j$ belongs to, i.e., $i:x_j \in \mathbf{x}^*_i$ to a hidden vector $\mathbf{e}^u_i \in \mathbb{R}^{d_{up}}$. Then, $\mathbf{e}^u_i$ is concatenated with the token embedding of $x_j$, i.e., $\mathbf{e}_{x_j}$ as an input to the encoder. That means the original encoder input $\mathbf{e}_{x_j}$ is replaced with $[\mathbf{e}_{x_j};\mathbf{e}^u_i]$.

For each generated summary token $y_{t-1}$, the token position $t-1$ is converted to a embedding vector $\mathbf{e}^s_{t-1} \in \mathbb{R}^{d_{sp}}$ through the summary token position embedding layer. After that, $\mathbf{e}^s_{t-1}$ is concatenated with $\mathbf{e}_{y_{t-1}}$ and then fed into the decoder, i.e., $\mathbf{e}_{y_{t-1}}$ is replaced with $[\mathbf{e}_{y_{t-1}} ; \mathbf{e}^s_{t-1}]$ as the decoder input. The SUFM embedding provides a word-level alignment  between the generated summary tokens and the dialogue utterances.

\smallskip
\noindent \textbf{SUFM Loss.}  Besides the word-level alignment, we also design an SUFM loss to provide the sentence-level alignment between the summary sentences and the input utterances, which further encourages the model to smoothly transit its focus on the input dialogue. As we defined before, the supporting utterances are the most informative ones for a summary sentence. We calculate the Jaccard similarity of the non-stop-word sets between a summary sentence and each dialogue utterance. We select the top-$N$ utterances as the supporting utterances of the summary sentence. The selection is based on the similarity scores and a similarity threshold. For each summary sentence, we will compute an SUFM loss. Using the $k$-th summary sentence as an example, we denote the current summary sentence as $\textit{CSS}_k$, the supporting utterances of the current summary sentence as $\textit{CSU}_k$, and the supporting utterances of previous summary sentences as $\textit{PSU}_k$. Our SUFM loss consists of two kinds of losses: $l_{\textit{CSU}_k}$ and $l_{\textit{PSU}_k}$. The $l_{\textit{CSU}_k}$ loss prompts our model to focus on the supporting utterances of the current summary sentence and is defined as:
\begin{align}
    l_{\textit{CSU}_k} = -\log(\frac{\sum_{t: y_t \in \textit{CSS}_k} \sum_{j:x_j \in \textit{CSU}_k} \bar{\alpha}^w_{t,j}}{\sum_{t: y_t \in \textit{CSS}_k} \sum_j \bar{\alpha}^w_{t,j}}) \text{,}
\end{align}
where $\bar{\alpha}^w_{t,j}$ is the final word-level attention score of each dialogue token. 

 The $l_{\textit{PSU}_k}$ loss encourages our model to ignore the supporting utterances of previous summary sentences when generating the current summary sentence. We define it as:
 \begin{align}
     l_{\textit{PSU}_k} = -\log(1 - \frac{\sum_{t: y_t \in \textit{CSS}_k} \sum_{j:x_j \in \textit{PSU}_k} \bar{\alpha}^w_{t,j}}{\sum_{t: y_t \in \textit{CSS}_k} \sum_j \bar{\alpha}^w_{t,j}}) \text{.}
 \end{align}

Then, we obtain $\lambda_1 l_{\textit{CSU}_k} + \lambda_2 l_{\textit{PSU}_k}$ as the SUFM loss of the $k$-th summary sentence, where $\lambda_1$ and $\lambda_2$ are hyperparameters. Consequently, we can get the total SUFM loss of a dialogue summary:
\begin{align}
    l_{SUFM} &= \sum^n_{k=1} (\lambda_1 l_{\textit{CSU}_k} + \lambda_2 l_{\textit{PSU}_k}) \text{,} \label{eq: l_sufm}
\end{align}
where $n$ is the number of sentences of $\mathbf{y}$.

\subsection{Fact Regularization}
In the fact regularization (FR) module, we first utilize a fact triplet extractor to extract fact triplets from the gold summary and then apply a regularization term in the training objective to help the model improve factual correctness when producing summaries. The fact extractor is based on the dependency parsing of each summary sentence. We engage the \textit{spacy}\footnote{\highlight{\url{https://pypi.org/project/spacy/2.1.0/}}} package to parse each summary sentence. After that, we extract the subject, the root verb, and the object tokens as a fact triplet (subject, verb, object) for the corresponding summary sentence. \highlight{Because the dependency parser is not absolutely precise, we sometimes cannot extract a complete subject-verb-object triplet from the given sentence. For such a case, we will skip this sentence to improve the accuracy of the extracted triplets from the summary.}
Based on the extracted fact triplets, we propose a regularization term which is inspired by the widely-known knowledge embedding approach TransE~\cite{bordes2013translating}. We assume that given the dialogue input, in the hidden representation space, summation of the subject and the verb should be close to the object as much as possible, i.e., $\mathbf{s}_\text{subject} + \mathbf{s}_\text{verb} \approx \mathbf{s}_\text{object}$. Consequently, based on the decoder hidden states $\mathbf{s}_i$ where $i=1,...,L_{\mathbf{y}}$, we introduce a TransE regularization for the extracted facts:
\begin{align}
    l_{FR} = \lambda_3 \sum_{k=1}^{m} ( 1 - \cos (\mathbf{s}_{subj_k} + \mathbf{s}_{verb_k}, \mathbf{s}_{obj_k}) ) \text{,} \label{eq: FR}
\end{align}
where $m$ is the total number of fact tuples extracted from the summary and $k$ indicates the $k$-th fact tuple. $\lambda_3$ is the weight hyperparameter. 
``$subj_k$'', ``$verb_k$'', and ``$obj_k$'' are indexes of the $k$-th fact tuple's subject, verb, and object.
We use the hidden states of the decoder to compute $l_{FR}$ since each summary token is generated based on its corresponding decoder hidden state. Therefore, each decoder's hidden state can be regarded as the dialogue-aware hidden representation of the generated summary token.

\subsection{Training}
Taking all the components into account, we use a joint loss to train our full model: $l = l_G + l_{SUFM} + l_{FR} \text{,}$
where $\lambda_1$, $\lambda_2$ in $l_{SUFM}$ and $\lambda_3$ in $l_{FR}$ are fine-tuned on the validation dataset. For our full  model, the SUFM embedding is included in the feature set $\mathcal{F}$ in $l_G$.

\section{Experiment Setup}
The implementation\footnote{The source code is released at \href{https://github.com/Chen-Wang-CUHK/DialSum-with-SUFM-and-FR}{https://github.com/Chen-Wang-CUHK/DialSum-with-SUFM-and-FR}} of our full model is based on the PyTorch~\cite{paszke2017automatic_pytorch} version of OpenNMT (ONMT) system~\cite{klein2017opennmt}. \highlight{Experiments of all neural-based models are repeated with three different random seeds for different parameter initialization.} The averaged results from these random seeds are reported. 

\subsection{Datasets}
Our experiments are conducted on two dialogue summarization datasets (the detailed statistics are shown in Table~\ref{tab: dataset_statistics}.):
\begin{itemize}[leftmargin=*]
    \item \textbf{SAMSum}~\cite{gliwa-etal-2019-samsum}. It is the first large scale, human-annotated, and public dialogue summarization dataset. There are 14,732 data examples for training, 818 for validation, and 819 for testing.
    
    \item \textbf{AVSD-SUM}\footnote{The preprocessed dataset is also released at \href{https://github.com/Chen-Wang-CUHK/DialSum-with-SUFM-and-FR}{https://github.com/Chen-Wang-CUHK/DialSum-with-SUFM-and-FR}}. \highlight{We generate this dataset from the existing video-grounded dialogue dataset \textit{DSTC7-AVSD}~\cite{DBLP:journals/corr/abs-1901-03461_DSTC7} to enrich the benchmarks of the dialogue summarization research area.} In \textit{DSTC7-AVSD}, a dialogue is generated by two human annotators based on a Charades video~\cite{DBLP:conf/eccv/SigurdssonVWFLG16}. After finishing the dialogue, one of the annotators summarizes the dialogue into a summary. We utilize the dialogue-summary pairs from \textit{DSTC7-AVSD} to build a new dialogue summarization dataset \textit{AVSD-SUM}. We filter out the data examples where the token number of the dialogue is less than 15 or the token number of the summary is less than 5. Finally,  we get 10,729 data examples. We randomly split 8,729 for training, 1,000 for validation, and 1,000 for testing.
\end{itemize}

\begin{table}[t]
    \centering
    \resizebox{0.7\columnwidth}{!}{
    \begin{tabular}{|c|c c c| c c c|}
    \hline
    \textbf{Dataset} & \textit{Train} & \textit{Valid} & \textit{Test} & \textit{Ave.U} &\textit{Ave.DL} & \textit{Ave.SL} \\
    \hline
    \textbf{SAMSum} & 14,732 & 818 & 819 & 11.1 & 126.7 & 23.5 \\
    \hline
    \textbf{AVSD-SUM} & 8,729 & 1,000 & 1,000 & 18.7 & 182.4 & 24.0 \\
    \hline
    \end{tabular}
    }
    \caption{Statistics of datasets. 
    ``\textit{Ave.U}'' is the averaged utterance number per dialogue. ``\textit{Ave.DL}'' means the averaged dialogue length (i.e., the number of tokens). ``\textit{Ave.SL}'' is the averaged summary length.}
    \label{tab: dataset_statistics}
\end{table}

\section{Implementation Details}
\noindent \textbf{Preprocessing}. We lowercase all the characters of dialogues and summaries. Then, we tokenize them into tokens using the \textit{spacy} package.

\smallskip
\noindent \textbf{Model}. The maximum size of the predefined vocabulary $\mathcal{V}$ is set as 50,000 and is shared between the encoder and decoder. If the vocabulary size of the dataset is smaller than 50,000, we include all the tokens. \highlight{Otherwise, we select 50,000 tokens with the highest token frequencies as the vocabulary. The summaries and the dialogues share the same vocabulary.} We set both the token embedding size $d_e$ and the hidden size $d$ as 300. The utterance position embedding size $d_{up}$ and the summary position embedding size $d_{sp}$ are set as $\lfloor V_f^{0.7}\rfloor $ where $V_f$ is the vocabulary size of the utterance position feature or the vocabulary size of the summary token position feature correspondingly.
The token embedding is initialized by the pre-trained \textit{GloVe}~\cite{DBLP:conf/emnlp/PenningtonSM14_glove} token embedding and then fine-tuned through training. \highlight{The coverage percentage of \textit{GloVe} on the vocabularies of SAMSum and AVSD-SUM are 91.9\% and 91.1\% respectively.} We initialize the hidden states of the encoder layers with zeros. In the training stage, we randomly initialize all the trainable parameters using a uniform distribution in $[-0.1, 0.1]$. 

\smallskip
\noindent \textbf{Loss}. When selecting the supporting utterances of each summary sentence, we set the selection number as 2 (i.e., $N$=2) and the Jaccard similarity threshold as 0.15. After fine-tuning on the validation datasets utilizing grid search on [0.1, 0.3, 1.0, 3.0], \highlight{we finally set $\lambda_1 = 0.3$ in Eq.(\ref{eq: l_sufm}), $\lambda_2 = 1.0$ in Eq.(\ref{eq: l_sufm}), $\lambda_3 = 0.3$ in Eq.(\ref{eq: FR}) for SAMSum dataset and $\lambda_1 = 0.1$, $\lambda_2 = 1.0$, $\lambda_3 = 0.3$ for AVSD-SUM dataset.}

\smallskip
\noindent \textbf{Training}. We set batch size as 32, initial learning rate as 0.001, and max gradient norm as 1.0. We set the dropout ratio $r$ as 0.2. Adam~\cite{DBLP:journals/corr/KingmaB14_adam} is used as our optimizer. 
The learning rate decays to half if the perplexity on the validation set stops decreasing. Early stopping is applied when training.

\smallskip
\noindent \textbf{Testing.} When testing, we set beam size as 5. The minimum and maximum decoding lengths are set as 15 and 100 separately. Repeated 2-grams are blocked~\cite{DBLP:conf/iclr/PaulusXS18}. We set $\alpha$ for length penalty~\cite{DBLP:journals/corr/WuSCLNMKCGMKSJL16} as 0.9 and $\beta$ for summary coverage penalty~\cite{DBLP:journals/corr/WuSCLNMKCGMKSJL16} as 5.

\subsection{Baseline Models and Evaluation Metrics}
For a comprehensive evaluation, we choose the following methods including both extractive and abstractive ones as our baselines:
\begin{itemize}[leftmargin=*]
    \item Extractive baselines: 
        \textbf{LONGEST-3}~\cite{gliwa-etal-2019-samsum} selects the longest 3 utterances as the summary. 
        \textbf{LexRank}~\cite{DBLP:journals/corr/abs-1109-2128_lexrank} extracts important utterances using a graph-based method. \textbf{BertSumExt}~\cite{liu-lapata-2019-text-bertsum} utilizes a BERT-based encoder to encode utterances and then performs sequence labeling process to extract important dialogue utterances.  \highlight{Following Gliwa et al.~\cite{gliwa-etal-2019-samsum}, we use these extractive baselines to extract utterances from the dialogue to compose a summary, where the maximum number of selected utterances is set to 3.}
        
    \item Abstractive baselines:
        \textbf{Fast Abs RL}~\cite{chen-bansal-2018-fast} first extracts salient utterances and then rewrites the extracted utterances abstractively. 
        \textbf{Fast Abs RL Enhanced}~\cite{gliwa-etal-2019-samsum} is an extension of \textit{Fast Abs RL} by appending the names of other speakers at the end of each utterance.
        \textbf{DynamicConv}~\cite{DBLP:conf/iclr/WuFBDA19} is a dynamic convolution based sequence-to-sequence framework.
        \textbf{ONMT-C.Transformer}~\cite{DBLP:conf/nips/VaswaniSPUJGKP17} is the OpenNMT~\cite{klein2017opennmt} (ONMT) implemented transformer framework with copy mechanism.
        \textbf{ONMT-PGNet}~\cite{see-etal-2017-get} is the OpenNMT implemented pointer generator.
        \textbf{ONMT-PGNet + GloVe} is an extension of \textit{ONMT-PGNet} by using GloVe~\cite{DBLP:conf/emnlp/PenningtonSM14_glove} to initialize the embedding matrix. 
        \textbf{BertSumExtAbs}~\cite{liu-lapata-2019-text-bertsum} fine-tunes a BERT encoder on the extractive summarization task first and then combines the BERT encoder with a transformer-based decoder to learn generating summaries abstractively. 
\end{itemize}

We also conduct ablation studies to evaluate the effectiveness of our newly-designed modules. We use \textbf{-SUFM} and \textbf{-FR} to respectively represent removing the SUFM module and the FR module from our full model.

When testing, all the ONMT-implemented models use the same testing setting as our model. For other models, we use the default testing settings but with the following adaptations. The beam size is set as 5. The minimum and maximum decoding lengths are set as 15 and 100, respectively. The only exception is that the \textit{Fast Abs RL} and \textit{Fast Abs RL Enhanced} cannot limit the minimum length of the generated summary since there is no such an option when inference. \highlight{But we also set the maximum number of selected utterances of these two methods as 3 in the extraction stage.}

For evaluation metrics, we employ the popular standard ROUGE~\cite{lin-2004-rouge} metric with stemming to evaluate all the methods. Similar to Gliwa et al.~\cite{gliwa-etal-2019-samsum}, we report the $F_1$ scores of ROUGE-1, ROUGE-2, and ROUGE-L. For simplicity, we use R-1, R-2, and R-L to represent these scores. All the rouge scores are computed through \textit{py-rouge}\footnote{\url{https://pypi.org/project/py-rouge/}}.

\begin{table*}[t]
\centering
\resizebox{0.9\textwidth}{!}{
\begin{tabular}{ | l | c c c | c c c |}
\hline
\multicolumn{1}{|c|}{\multirow{2}{*}{\textbf{Model}}} & \multicolumn{3}{c|}{\textbf{SAMSum}} & \multicolumn{3}{c|}{\textbf{AVSD-SUM}} \\ \cline{2-7}
\multicolumn{1}{|c|}{}   & R-1    & R-2    & R-L & R-1  & R-2    & R-L \\
\hline \hline
LONGEST-3
& $31.60 \pm .00^*$ & $9.91 \pm .00^*$ & $27.26 \pm .00^*$
& $34.31 \pm .00^* $ & $14.79 \pm .00^*$ & $28.97 \pm .00^*$ \\

LexRank 
& $23.05 \pm .00^*$ & $4.60 \pm .00^*$ & $20.45 \pm .00^*$
& $31.50 \pm .00^* $ & $12.68 \pm .00^*$ & $26.17 \pm .00^*$ \\

BertSumExt
& $37.00 \pm .14^* $ & $13.08 \pm .10^* $ & $34.52 \pm .14^* $
& $45.29 \pm .12^* $ & $21.60 \pm .10^* $ & $40.74 \pm .09^* $ \\
\hline

Fast Abs RL
& $40.14 \pm .34^*$ & $16.54 \pm .23^*$ & $ 38.38 \pm .40^*$
& $47.42 \pm .38^*$ & $22.33 \pm .07^*$ & $43.17 \pm .29^*$ \\ 

Fast Abs RL Enhanced
& $40.66 \pm .72^*$ & $16.78 \pm .59^*$ & $38.34 \pm .96^*$
& N/A & N/A & N/A \\

DynamicConv
& $35.95 \pm .77^*$ & $14.08 \pm .55^*$ & $33.23 \pm .44^*$
& $51.65 \pm .42^*$ & $25.14 \pm .28^*$ & $46.74 \pm .39^*$ \\

ONMT-C.Transformer
& $37.69 \pm .37^*$ & $11.29 \pm .22^*$ & $34.05 \pm .15^*$
& $48.54 \pm .31^*$ & $20.79 \pm .15^*$ & $42.76 \pm .27^*$ \\

ONMT-PGNet
& $40.94 \pm .10^*$ & $16.71 \pm .09^*$ & $37.86 \pm .14^*$
& $53.35 \pm .39$ & $27.94 \pm .27^*$ & $48.36 \pm .30^*$ \\ 

ONMT-PGNet + GloVe 
& $42.28 \pm .09^*$ & $17.90 \pm .34^*$ & $39.11 \pm .22^*$
& $53.65 \pm .07^*$ & $28.26 \pm .11^*$ & $48.82 \pm .18^*$ \\ 

BertSumExtAbs 
& $40.75 \pm .42^*$ & $18.01 \pm .29^*$ & $37.99 \pm .32^*$
& $53.78 \pm .36$ & $27.32 \pm .28^*$ & $48.91 \pm .26^*$ \\
\hline
Our Model
& \textbf{42.85 $\pm$ .17} & \textbf{18.59 $\pm$ .24} & \textbf{39.84 $\pm$ .13}
& \textbf{54.38 $\pm$ .11} & \textbf{28.95 $\pm$ .05} & \textbf{49.54 $\pm$ .10} \\ 
\hline

-SUFM 
& $42.45 \pm .27$ & $18.17 \pm .31$ & $39.45 \pm .32$
& $54.17 \pm .15$ & $28.56 \pm .28$ & $49.32 \pm .13$ \\ 

-FR 
& $42.72 \pm .08$ & $18.33 \pm .06$ & $39.73 \pm .06$
& $53.94 \pm .24$ & $28.52 \pm .31$ & $49.11 \pm .31$ \\ 
\hline
\end{tabular}
}
\caption{
ROUGE scores of all the models on SAMSum and AVSD-SUM datasets. The best results are bold. Note that the implementation of our model is based on the implementation of ONMT-PGNet+GloVe.
The ``*'' indicates our model significantly (paired t-test, p $<$ 0.05) outperforms the corresponding baseline.
}
\label{table: main-results}
\vspace{-0.1in}
\end{table*}

\section{Results and Analysis}
\subsection{Main Results}
The ROUGE scores of all the models on SAMSum and AVSD-SUM datasets are displayed in Table~\ref{table: main-results}. We find that our model consistently outperforms all the baselines on both datasets, which indicates the superiority of our model.
We also observe that the \textit{ONMT-PGNet-GloVe} outperforms the BERT-based models (i.e., \textit{BertSumExt} and \textit{BertSumExtAbs}) on most cases. This is the main reason that we choose \textit{ONMT-PGNet-GloVe} as the implementation basis of our model.
The \textit{Fast Abs RL Enhanced} method cannot be applied to the AVSD-SUM dataset because there is no information about the speakers' names in this dataset. Besides, we also note the decreasing of the ROUGE scores after removing either the SUFM module or the FR module, which shows the effectiveness of these modules. We also see that there is a large performance gap between the extractive methods like BertSumExt and the abstractive methods like BertSumExtAbs, which is much different from the news summarization task where BertSumExt achieves higher ROUGE scores than BertSumExtAbs. This means dialogue summaries are much more abstractive than news summaries and extracting several utterances from the dialogue as the summary is not effective.

\subsection{SUFM Analysis}
\begin{figure}[t]
\centering
\includegraphics[width=0.7\columnwidth]{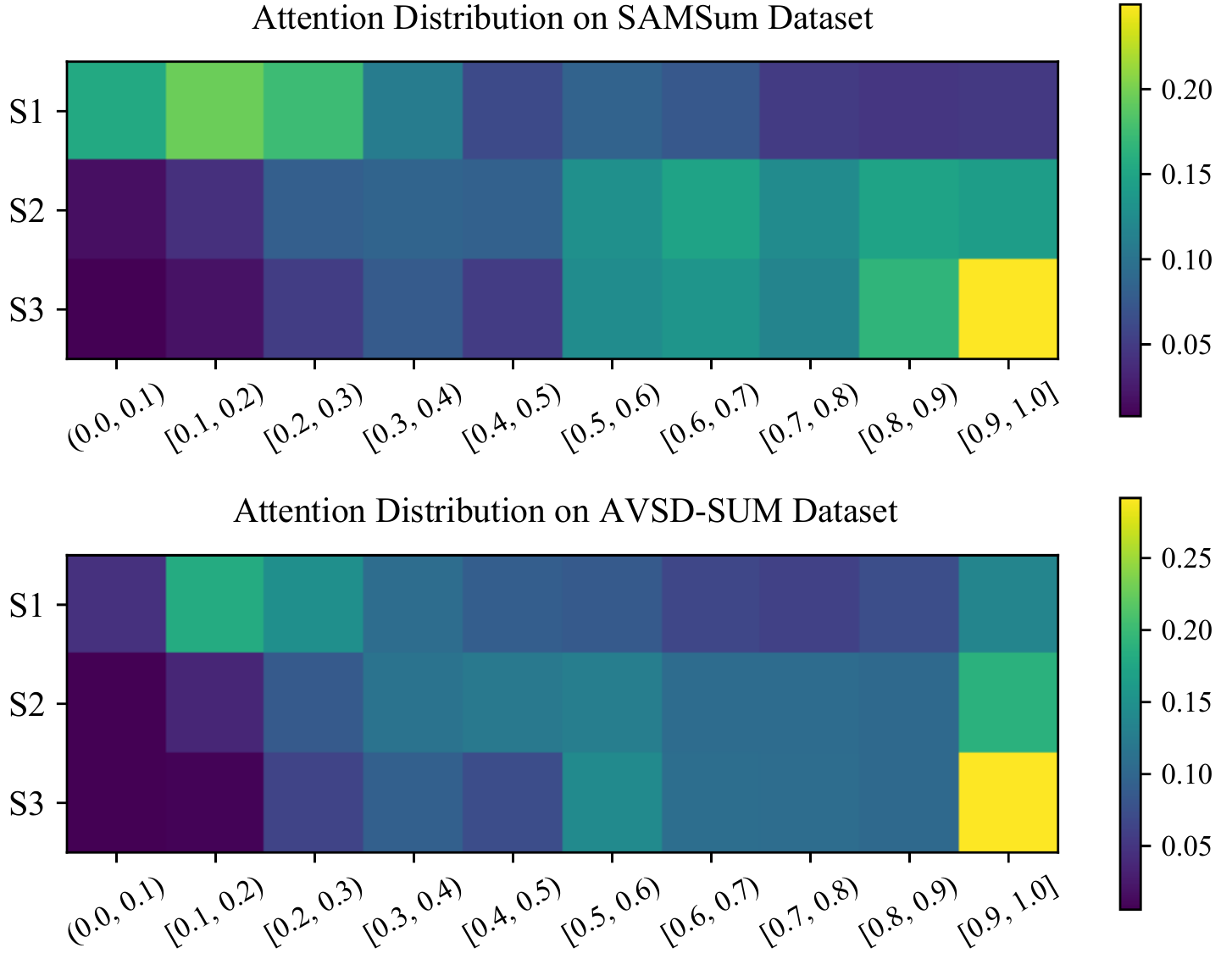}
\vspace{-0.1in}
\caption{The averaged attention distribution of our full model on SAMSum and AVSD-SUM testing datasets. ``S$i$'' is the $i$-th generated summary sentence. The horizontal axis indicates the relative utterance position e.g., ``[0.1, 0.2)'' means the range of the first 10\%-20\% utterances.}
\label{figure: sufm_attn_distribution}
\vspace{-0.1in}
\end{figure}

We also study whether our full model can smoothly transit its focus from the start to the end of the dialogue when generating a summary. To evaluate such capacity, we display the averaged attention distribution of each generated summary sentence over the relative utterance position range in Figure~\ref{figure: sufm_attn_distribution}.

From this figure, we can find that the first generated summary sentence S1 of both datasets mainly focuses on the former part of the dialogue. The second generated sentence S2 of the SAMSum dataset mostly focuses on the 50\%-100\% part of the dialogue. The generated S2 of AVSD-SUM dataset focuses on 30\%-60\% part and 90\%-100\% part of the dialogue. The third generated sentence S3 of both datasets mostly focuses on the ending part of the dialogue. Therefore, we can observe an explicit attention transition from the start to the end of the dialogue, which manifests our full model can effectively capture the supporting utterance flow information. We also note that the generated S2 of both datasets obtains a high attention score on the 90\%-100\% range. The reason is that most of the \highlight{generated summaries} of these two testing datasets (over 77\% on SAMSum, over 66\% on AVSD) only contain two sentences, which means S2 is the last sentence of most summaries. Thus, S2 may pay high attention to the ending part of the dialogue.

\begin{table}[t]
\centering
\resizebox{0.7\columnwidth}{!}{
\begin{tabular}{ | l | c | c |}
\hline
\multicolumn{1}{|c|}{\multirow{2}{*}{\textbf{Model}}} & \multicolumn{1}{c|}{\textbf{SAMSum}} & \multicolumn{1}{c|}{\textbf{AVSD-SUM}} \\ \cline{2-3}
& $F_1$ & $F_1$\\
\hline
LONGEST-3
& $6.10 \pm .00^*$ 
& $13.20 \pm .00^*$ \\

LexRank
& $3.00 \pm .00^*$
& $13.80 \pm .00^*$\\

BertSumExt 
& $8.07 \pm 0.21^*$ 
& $23.13 \pm .32^*$ \\
\hline

Fast Abs RL
& $21.10 \pm 1.44$ 
& $26.77 \pm .35^*$ \\

Fast Abs RL Enhanced 
& $21.47 \pm 1.58$ 
& N/A \\

DynamicConv 
& $18.77 \pm .38^*$ 
& $34.93 \pm .96$ \\

ONMT-C.Transformer
& $15.23 \pm .15^*$ 
& $28.40 \pm 1.21^*$ \\

ONMT-PGNet 
& $19.77 \pm .85$ 
& $33.20 \pm .26^*$ \\

ONMT-PGNet + GloVe 
& $21.73 \pm 1.27$ 
& $33.90 \pm .46^*$ \\

BertSumExtAbs 
& \underline{21.80 $\pm$ .17} 
& \textbf{35.73 $\pm$ .51} \\
\hline

Our Model
& \textbf{22.30 $\pm$ .90} 
& \underline{35.47 $\pm$ .47} \\
\hline

-SUFM 
& $21.53 \pm .50$ 
& $34.47 \pm .61$ \\

-FR
& $21.40 \pm .30$ 
& $34.50 \pm .90$ \\
\hline
\end{tabular}
}
\caption{The $F_1$ scores of fact triplet matching. The best results are bold and the second-best results are underlined. The ``*'' indicates our model significantly (paired t-test, p $<$ 0.05) outperforms the corresponding baseline.
}
\label{table: fact_triplet_matching}
\vspace{-0.1in}
\end{table}

\subsection{Fact Triplet Matching}
In this section, we evaluate how many fact triplets extracted from the \highlight{generated summaries} are matched with gold fact triplets (i.e., precision) and how many gold triplets are covered by the predicted fact triplets (i.e., recall). To achieve such a goal, we compute the micro-averaged $F_1$ score between the predicted fact triplets from the \highlight{generated summaries} and the gold fact triplets from the gold summaries. \highlight{When calculating the score, we regard two fact triplets are (partially) matched if the number of the overlapped components is at least two.} The results are listed in Table~\ref{table: fact_triplet_matching}.

From this table, we find that our model outperforms almost all the baselines on both datasets, which indicates that our model is more effective in predicting important facts from the dialogue. We also note that our model gets slightly lower $F_1$ scores than BertSumExtAbs on AVSD-SUM. One potential reason is that BertSumExtAbs utilizes the pre-trained encoder BERT~\cite{devlin-etal-2019-bert} in its encoding process, but we do not incorporate it. We leave incorporating BERT into our model as future work. From the table, we also see that after removing the fact regularization module (i.e., -FR), the $F_1$ score drops as we anticipated. Another interesting finding is that our SUFM module is also helpful in predicting more accurate facts. The possible reason is that it helps the model transit the focus from the beginning to the end of the dialogue and improves the chance to predict the crucial facts that are located in different parts of the dialogue. 

\subsection{Human Evaluation}

We randomly select 50 examples from SAMSum to conduct human evaluation. The \highlight{generated summaries} from humans (i.e., the gold summaries), ONMT-PGNet + GloVe, BertSumExtAbs, and our model are evaluated by three human raters. Each rater reads the dialogue of each example and then scores the \highlight{generated summaries} with a rating scale from 1 (worst) to 5 (best) on the following three aspects: (1) \textbf{Grammaticality} assesses how fluent and grammatical the summary is. (2) \textbf{Informativeness} measures how much salient information the summary contains. It can also reflect how many important facts are correctly predicted in the generated summary. (3) \textbf{Coherence} evaluates whether the summary presents content in a coherent order.
The averaged results are listed in Table~\ref{table: human_evaluation}. We note that a large margin exists between the gold summaries and model \highlight{generated summaries} on these three aspects, which indicates there is still a large distance to produce grammatical, informative, and coherent summaries automatically. We also find that based on the human evaluation, our model still consistently outperforms these two state-of-the-art baselines. 

\begin{table}[t]
\centering
\resizebox{0.7\columnwidth}{!}{
\begin{tabular}{ | l | c | c | c |}
\hline
\textbf{Model} & Grammaticality & Informativeness  & Coherence \\
\hline
ONMT-PGNet + GloVe 
& 3.83
& 2.76
& 3.32
\\

BertSumExtAbs 
& 3.96 
& 2.65
& 3.37
\\

Our Model
& \textbf{4.08} 
& \textbf{2.95}
& \textbf{3.55}
\\

\hline
Gold
& 4.65
& 4.54
& 4.59
\\
\hline
\end{tabular}
}
\caption{The human evaluation results.}
\label{table: human_evaluation}
\vspace{-0.1in}
\end{table}

\section{Conclusion}
In this paper, we introduce a new dialogue summarization model that incorporates two newly introduced modules: the SUFM module and the FR module. The SUFM module helps our model smoothly transit its focus from the beginning to the end of the dialogue. The FR module helps our model improve factual correctness when generating summaries. Besides a new model, we also introduce a new dataset to enrich the benchmarks of this community. Extensive experiments show the effectiveness of our proposed model. One meaningful future direction is to investigate how to effectively incorporate the pre-trained encoder like BERT into our model and fine-tune it.

\section*{Acknowledgements}
The work described in this paper was partially supported by the National Key Research and Development Program of China (No. 2018AAA0100204) and the Research Grants Council of the Hong Kong Special Administrative Region, China (CUHK 2410021, Research Impact Fund, R5034-18).


\bibliography{mybibfile}


\end{document}